\documentclass[letterpaper]{article} % DO NOT CHANGE THIS
\usepackage{aaai2026}  % DO NOT CHANGE THIS
\nocopyright
\usepackage{times}  % DO NOT CHANGE THIS
\usepackage{helvet}  % DO NOT CHANGE THIS
\usepackage{courier}  % DO NOT CHANGE THIS
\usepackage[hyphens]{url}  % DO NOT CHANGE THIS
\usepackage{graphicx} % DO NOT CHANGE THIS
\usepackage{natbib}  % DO NOT CHANGE THIS AND DO NOT ADD ANY OPTIONS TO IT
\usepackage{caption} % DO NOT CHANGE THIS AND DO NOT ADD ANY OPTIONS TO IT
\usepackage{algorithm}
\usepackage{algorithmic}

\usepackage{newfloat}
\usepackage{listings}
\DeclareCaptionStyle{ruled}{labelfont=normalfont,labelsep=colon,strut=off} % DO NOT CHANGE THIS
\floatstyle{ruled}
\newfloat{listing}{tb}{lst}{}
\floatname{listing}{Listing}
\usepackage{amssymb}
\usepackage{amsmath}
\usepackage{booktabs}
\usepackage{tabularx}
\usepackage{multirow}
\usepackage{array}
\usepackage{adjustbox}
\usepackage{bm}

\usepackage{xcolor}
\usepackage{pifont}
\definecolor{forestgreen}{rgb}{0.13, 0.55, 0.13}
\definecolor{triangle}{rgb}{0.9, 0.55, 0.2}
\definecolor{circle}{rgb}{0.45, 0.6, 0.85}
\newcommand{\cmark}{{\color{forestgreen} \ding{51}}}%
\newcommand{\xmark}{{\color{red} \ding{55}}}%
\newcommand{\circlemark}{{\color{triangle} \ding{108}}}%
\newcommand{\trimark}{{\color{circle} \ding{115}}}%

\usepackage{tikz}
\usepackage{pgfplots}
\pgfplotsset{compat=1.18} % Use a recent version
\usepgfplotslibrary{groupplots} % For subplots
\usetikzlibrary{plotmarks}      % For more marker styles (like pentagon)
\definecolor{dhcolor}{RGB}{0,114,178} % StaticDH:    #0072B2 (Color Blind Safe Blue)
\definecolor{ddhcolor}{RGB}{230,159,0} % DynamicDH:   #E69F00 (Color Blind Safe Orange)
\definecolor{tscolor}{RGB}{0,158,115} % Transformer: #009E73 (Color Blind Safe Green)
\definecolor{stcolor}{RGB}{213,94,0} % SurvTrace:   #D55E00 (Color Blind Safe Vermillion)
\definecolor{dscolor}{RGB}{204,121,167} % DeepSurv:    #CC79A7 (Color Blind Safe Reddish Purple)
\definecolor{cphcolor}{RGB}{86,180,233} % CoxPH:       #56B4E9 (Color Blind Safe Sky Blue)
\pgfplotsset{
    modelplot/.style={
        smooth,
        line width=3pt,      % Thicker than 'very thick'
        opacity=0.8,         % Slightly transparent
        mark=*,
        mark size=1.5pt
    }
}

\newcommand{{\tracer}}{{\textsc{TraCeR}}}

\title{{\tracer}: Transformer-Based Competing Risk Analysis with Longitudinal Covariates}
\author {
    Maxmillan Ries\textsuperscript{\rm 1},
    Sohan Seth\textsuperscript{\rm 1},
}
\affiliations {
    \textsuperscript{\rm 1}University of Edinburgh\\
    M.V.Ries@sms.ed.ac.uk, sohan.seth@ed.ac.uk
}
\begin{document}

\maketitle

\begin{abstract}
Survival analysis is a critical tool for modeling time-to-event data. Recent deep learning-based models have reduced various modeling assumptions including proportional hazard and linearity. However, a persistent challenge remains in incorporating longitudinal covariates, with prior work largely focusing on cross-sectional features, and in assessing calibration of these models, with research primarily focusing on discrimination during evaluation. We introduce {\tracer}, a transformer-based survival analysis framework for incorporating longitudinal covariates. Based on a factorized self-attention architecture, {\tracer} estimates the hazard function from a sequence of measurements, naturally capturing temporal covariate interactions without assumptions about the underlying data-generating process. The framework is inherently designed to handle censored data and competing events. Experiments on multiple real-world datasets demonstrate that {\tracer} achieves substantial and statistically significant performance improvements over state-of-the-art methods. Furthermore, our evaluation extends beyond discrimination metrics and assesses model calibration, addressing a key oversight in literature.
\end{abstract}


\section{Introduction}
Time-to-event analysis, or survival analysis, is a statistical framework for modeling the duration until an event occurs \cite{survival-def}. It aims to understand the relationship between the timing of an event (e.g., equipment failure or disease onset) and a set of predictive covariates (e.g., telemetry or lab results). The focus of this analysis is the \textit{hazard function}, which quantifies the instantaneous risk of an event at a specific time, given that the event has not yet occurred. This framework is a powerful tool for decision making in fields ranging from medicine, where it guides treatment plans, to finance, where it can model credit card default \cite{ddhit}.
However, applying survival analysis to real-world data introduces two significant challenges that can lead to biased models if ignored. First, data is often \textit{censored}: the true event time is masked because a subject is lost to follow-up or the observation period ends, i.e., we only know that the subject was event-free up to a certain point in time. Second, subjects are often exposed to \textit{competing risks}, where the occurrence of one event (e.g., patient death) precludes the occurrence of another (e.g., onset of cancer). The presence of competing risks introduces a critical distinction from censoring. While a censored subject is assumed to remain ``at risk'' for the event of interest, a subject who experiences a competing event is not, and therefore, not addressing for competing risk might lead to overestimation of risk \cite{competing_risk_censoring}. 
This distinction has implications for the application of classical survival models. For instance, in the widely used Cox Proportional Hazards (CoxPH) model, competing risks are treated as censored observations \cite{Cox-competing-as-censoring}. This simplification is statistically inconsistent with the non-informative censoring assumption. %, as it fails to remove the subject from the at-risk population. This leads to biased estimates of the event-specific hazard.
Furthermore, classical models are frequently constrained by other fundamental assumptions, e.g., the CoxPH model operates under a strict ``proportional hazard'' assumption, requiring the ratio of hazard rates between subjects to be constant over time \cite{mtlsurvival}. Moreover, such models are not inherently designed to utilize longitudinal data (e.g., routine lab tests) and often require experts to derive a cross-sectional summary vector.

\begin{table*}[t]
  \small
  \renewcommand{\arraystretch}{0.9}   % tighter rows
  \setlength{\tabcolsep}{3pt}         % tighter columns
  \centering

  \begin{tabular}{lcccccccccc}
    \hline
    \textbf{Model} & \textbf{Static Cov.} & \textbf{Long. Cov.}
      & \textbf{Comp. Events} & \textbf{Censor.} & \textbf{Miss. Cov.}
      & \textbf{Non-P.H.\textsuperscript{1}} & \textbf{Discr.\textsuperscript{2}} & \textbf{Cali.\textsuperscript{2}} & \textbf{Output} & \textbf{Core Arch.} \\
    \hline
    CoxPH            & \cmark & \xmark   & \trimark  & \cmark  & \xmark
                     & \xmark & \cmark & \cmark           & Log-risk      & Linear Model   \\
    DeepSurv         & \cmark & \xmark   & \xmark     & \cmark  & \xmark
                     & \xmark & \cmark & \xmark           & Log-risk  & MLP            \\
    DeepHit          & \cmark & \xmark   & \cmark & \cmark  & \xmark
                     & \cmark & \cmark & \xmark           & PMF         & MLP            \\
    DynamicDeepHit   & \cmark & \cmark & \cmark & \cmark & \cmark
                     & \cmark & \cmark & \cmark           & PMF         & RNN           \\
    MMT$^*$ \shortcite{MMT} & \cmark & \cmark & \xmark & \cmark & \xmark & \cmark & \cmark & \xmark & Hazard & Transformer \\
    N-MTLR$^*$ \shortcite{mtlsurvival}           & \cmark & \xmark   & \xmark     & \cmark  & \xmark
                     & \cmark & \cmark & \circlemark           & Hazard         & MLP            \\
    SurvFormer$^*$ \shortcite{SurvFormer}        & \cmark & \xmark   & \xmark & \cmark  & \xmark
                     & \cmark & \cmark & \xmark           & Hazard      & Transformer    \\
    SurvTRACE        & \cmark & \xmark   & \cmark & \cmark  & \xmark
                     & \cmark & \cmark & \xmark           & Hazard      & Transformer    \\
    SurvTimeSurvival & \cmark & \cmark & \xmark     & \cmark  & \xmark
                     & \cmark & \cmark & \circlemark           & Hazard      & Transformer    \\
    TBDSA$^*$ \shortcite{TBDSA} & \cmark & \xmark & \xmark & \cmark & \xmark & \cmark & \cmark & \cmark & Hazard & Transformer \\
    {\tracer}        & \cmark & \cmark & \cmark & \cmark  & \cmark
                     & \cmark & \cmark & \cmark           & Hazard      & Transformer    \\
    \hline
  \end{tabular}

  \caption{Summary of prior work. \trimark: With extensions. \circlemark: Partially. \textsuperscript{1}: Proportional Hazard. $^*$Not included in comparison. \textsuperscript{1} Discr/Cali reported}
      \label{tab:related-works}
\end{table*}
To overcome these limitations, recent research employed recurrent neural architectures (RNNs), such as Gated Recurrent Units (GRUs) and Long-Short Term Memory (LSTM) networks \cite{RNNs}. These data-driven models allow the relaxation of strong parametric assumptions and directly allow the inclusion of longitudinal measurements, allowing the temporal relationship between subject history and risk to be captured. While a significant improvement over classical models, the sequential and summarization nature of RNNs struggle with capturing long-term dependencies \cite{RNN-long-term-dependencies}. 
More recently, the discovery of the Transformer architecture has presented a compelling solution, demonstrating superior performance in capturing long-range dependencies and complex interactions to RNNs. Models such as SurvTRACE first demonstrated the capability of self-attention to learn complex relationships in cross-sectional data \cite{SurvTRACE}. Subsequent work, such as SurvTimeSurvival, extended this work to longitudinal settings by employing a transformer to summarize temporal information, and subsequently feeding the representation into SurvTRACE \cite{survtimesurvival}. 

\subsubsection{Contribution} This paper presents a Transformer-based architecture, {\tracer}, for survival analysis with longitudinal covariates. We extend previous work applying the Transformer architecture to (predominantly) cross-sectional data, to a longitudinal setting. To robustly handle missing data that is ubiquitous in clinical applications, {\tracer} first employs a \emph{covariate missingness} gate that uses a time-decay function to modulate feature importance based on the time elapsed since the last observation \cite{timedecaymissing}. The processed inputs are then fed into a \textit{factorized attention} mechanism that operates across both the temporal and covariate axes \cite{factorizedattention}. This allows the model to explicitly capture the evolving influence of covariates over time. Rather than using only the final hidden state, {\tracer} then generates a single context vector through a learned summarization of the entire sequence of temporal outputs. This vector, which represents the subject's history, is finally passed to \emph{cause-specific subnetworks} to estimate the discrete-time hazards for each potential event. 

To demonstrate the effectiveness of our method, we compare {\tracer}'s performance against competing prior works on four challenging longitudinal datasets derived from the \textsc{eICU} \shortcite{eICU}, \textsc{MIMIC-IV} \shortcite{mimiciv}, \textsc{PBC2} \shortcite{pbc2}, and \textsc{MIMIC-III} \shortcite{mimiciii}. These cohorts feature competing risks and, in the \textsc{eICU} dataset, a high degree of censoring. We evaluate performance on two key aspects, \textit{discrimination} and \textit{calibration}. Our results show that {\tracer} achieves state-of-the-art performance on both fronts. Critically, we also discuss how these two aspects of performance are often conflated in prior work. Our deeper analysis reveals that {\tracer} provides not only superior discrimination but also significantly better-calibrated risk estimates, a crucial factor for clinical trust and utility.

\section{Related Work}
The most widely used statistical model in medical research and clinical survival analysis is the Kaplan-Meier estimation. While fundamental for estimating survival functions from a population, it does not incorporate subject-specific covariates into its estimation \cite{kaplan}. To address this, the semi-parametric CoxPH model has become a cornerstone of clinical trials and observational studies by allowing the inclusion of a static set of subject covariates. Though powerful, its utility is limited by the proportional hazard assumption, and many machine learning and deep learning methods have been developed to relax these assumptions and improve predictive power. A comprehensive overview of prior machine learning methods for survival analysis can be found in \cite{related-work-ml}.

Deep learning first addressed these limitations in the context of static, cross-sectional data with models like DeepSurv \cite{deepsurv} and DeepHit \cite{deephit}. DeepSurv replaces the linear risk function of the CoxPH model with a deep neural network, enabling it to learn complex, non-linear interactions between covariates and risk, under the assumption of proportional hazard. DeepHit reframes survival analysis as a multi-task classification problem, directly learning the joint probability distribution over event times and competing risks. Though powerful, these models were designed for a static cross-sectional snapshot of subject data, limiting their application where longitudinal measurements are available.

Building on these foundational deep learning models, follow-up work began integrating longitudinal measurements. DynamicDeepHit extends the DeepHit architecture by employing a shared RNN, such as an LSTM or GRU, to encode the evolution of subject covariates over time \cite{ddhit}. The subject history is processed sequentially and compressed into a context vector. This vector serves as a summary of the past, and is then passed to cause-specific subnetworks to derive first-event competing risk predictions. While a significant improvement over cross-sectional input models, this approach inherits the fundamental limitations of RNNs. The sequential nature of RNNs poses a challenge for modeling long-range dependencies \cite{RNN-long-term-dependencies}.

More recently, the Transformer architecture \cite{transformer}, with its self-attention mechanism, has emerged as a powerful alternative for modeling temporal and covariate interactions without the sequential constraints of RNNs. Its application in survival analysis was recently explored by models such as SurvTRACE \shortcite{SurvTRACE}, which leverages self-attention over a set of cross-sectional embeddings to learn high-order covariate interactions. The resulting representation is then used to predict discrete-time hazards. While this approach demonstrated superior and more interpretable modeling of covariate interactions, its design is fundamentally limited to cross-sectional data.
Expanding SurvTRACE, SurvTimeSurvival employs an encoder to distill a subject's multivariate time-series into a single summary vector \shortcite{survtimesurvival}. This summary vector is then fed into the core SurvTRACE model. Although extending SurvTRACE to longitudinal measurements, this two-stage approach separates the processing of temporal dynamics from cross-covariate interactions. By first creating a static summary of subject history, the model is unable to explicitly learn how the influence of covariates vary over time. Moreover, SurvTimeSurvival was only designed and tested on single-event analysis, and does not address competing risks, a requirement in clinical survival analysis.

While recent models have begun leveraging deep learning for survival analysis, critical gaps remain. RNN-based approaches struggle with long-range dependencies and implicitly model covariate interactions. Current Transformer-based methods are either restricted to cross-sectional data or enforce a separation between temporal and feature interactions. This separation not only prevents the model from capturing the evolving influence between risk factors, but also ignores the challenge of handling the informative missingness inherent in longitudinal data. Our work, {\tracer}, addresses these modeling limitations. Furthermore, we tackle the partial evaluation gap in prior work (Table~\ref{tab:related-works}) by conducting a comprehensive analysis of both discrimination and calibration metrics.

\section{Problem Formulation}
In the medical domain, longitudinal survival data for a cohort of $N$ subjects can be formally represented as a set $\mathcal{D} = \{(\mathcal{X}_i, t_i, e_i)\}_{i=1}^{N}$. For each subject $i$, $\mathcal{X}_i$ represents the subject's covariate history, a time-ordered sequence of $D$ covariates $\mathcal{X}_i = (\mathbf{x}_{i,t^i_1}, \mathbf{x}_{i,t^i_2} \cdots \mathbf{x}_{i,t^{i}_{m_i}})$ with $t^{i}_{m_i} < t_i$. This trajectory can contain missing values and accommodates for both static features measured once (e.g., age at admission, ethnicity) and dynamic features measured repeatedly over time (e.g., lab results, vital signs). The tuple $(t_i, e_i)$ captures the event outcome, where $t_i$ is the observed event time. The event indicator $e_i = k$ where $k \in \{0, 1, \cdots, K\}$ specifies which of the $K$ competing events occurred, while $e_i = 0$ denotes that the subject was censored at time $t_i$. Supp. Mat. A1 provides an illustration of clinical survival data, with subject trajectories aligned on a common timeline. 

Real-world clinical data is often sampled irregularly. To build models that can process such time-series, we discretize time into a set of intervals, $j = 1, \cdots, J$. This process transforms the continuous outcome $t_i$ into a discrete event interval $T_i$. Our goal is to predict an subject's future risk conditioned on their discretized longitudinal covariates history $\mathbf{X}_i$. Specifically, using a subject's history $\mathbf{X}_i \in \mathbb{R}^{\tau_i \times D}$, recorded up to a landmark interval $\tau_i$, we want to estimate the probability of them experiencing each of the $K$ competing events in subsequent intervals.
This is achieved by modeling one of two fundamental quantities: the probability mass function (PMF) or the cause-specific hazard function. The latter, which is used by \textsc{TraCeR}, offers greater clinical interpretability by modeling the instantaneous risk of a specific event at a given time, and a more principled statistical structure. In our discrete-time setting, the cause-specific hazard $\lambda^k_{j}(\mathbf{X}_i)$ is the conditional probability of subject $i$ experiencing event $k$ in interval $j$, given they have not experienced any event up to the beginning of that interval, and conditioned on their covariate history $\mathbf{X}_i$ \cite{bible}, i.e.,
{\small
\begin{equation}
\label{eq:hazard_softmax}
\begin{aligned}
\lambda_{j}^{k}(\mathbf{X}_i)
&= P(T_i = j,\; e_i = k \mid T_i \ge j,\; \mathbf{X}_i) \\
&= \frac{\exp\!\big(f_{\theta}(\mathbf{X}_i)_{j,k}\big)}
       {1 + \sum_{m=1}^{K} \exp\!\big(f_{\theta}(\mathbf{X}_i)_{j,m}\big)} ,
\end{aligned}
\end{equation}
}
where $f_\theta(\mathbf{\hat{X}}_i)_{j,k}$ is the model’s prediction for subject $i$ in interval $j$ and cause $k$. The resulting hazards are used to compute the overall survival probability $S_j(\mathbf{X}_i)$, i.e., the likelihood of remaining event-free, as 
{\small
\begin{equation}
S_{j}(\mathbf{X}_i) = \prod_{l=1}^{j} \left( 1 - \sum_{k=1}^{K} \lambda_{l}^k(\mathbf{X}_i) \right).
\end{equation}
} 
Most critically for competing risks, we can compute the Cumulative Incidence Function (CIF), $F_j^k(\mathbf{X}_i)$, for each event type, which gives the probability of event $k$ occurring by or within interval $j$,
{\small
\begin{equation}
F_j^k(\mathbf{X}_i) = P(T_i \le j, e_i=k \,|\, \mathbf{X}_i) = \sum_{l=1}^{j} \lambda_l^k(\mathbf{X}_i) \cdot S_{l-1}(\mathbf{X}_i).
\end{equation}
}
Note that, we learn the discretized hazard function, survival function and CIF for all future time bin intervals $j > \tau_i$.

\begin{figure*}[!t]
\centering
\includegraphics[
    width=1.0\linewidth
]{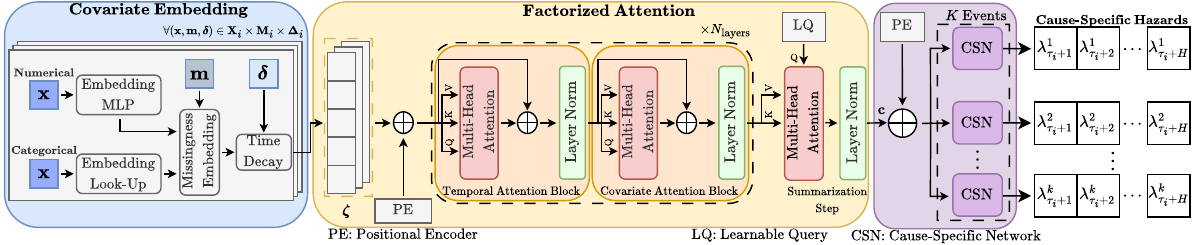}
\caption{{\tracer} Architecture. The raw sequence of numerical and discrete covariates are encoded per time-step through two embedding paths, before missingness and time-decay is applied. The resulting embeddings $(\tau_i, D, d_{\text{emb}})$ are fed into factorized attention layers. The resulting combinatorial embedding is summarized through a learnable query and passed onto $K$ cause-specific subnetworks. Each subnetwork produces a sequence of $\lambda_j^k(\mathbf{X}_i)$ hazards.}
\label{fig:arch}
\end{figure*}

\section{Method}
\subsection{Network Architecture}
The architecture of {\tracer}, illustrated in Figure~\ref{fig:arch}, is composed of three main components that map a subject's longitudinal history to a sequence of cause-specific hazards. The model begins with the \textit{Time-Aware Covariate Embedding} module. This module takes as input three tensors representing the subject's history up to a landmark interval $\tau_i$: i) the discretized longitudinal covariates $\mathbf{X}_i$; ii) a binary missingness mask $\mathbf{M}_i$ indicating observed versus imputed values; and iii) a time-delta tensor $\mathbf{\Delta}_i$ that records the time elapsed since the last non-imputed observation for each covariate. The resulting time-aware embeddings are then passed to the \textit{Factorized Attention Encoder}, which models temporal and covariate interactions. Finally, the encoded representation is processed by a set of \textit{Cause-Specific Hazard Subnetworks} to produce the final discrete hazard predictions.

\subsubsection{Time-Aware Covariate Embedding}
The time aware embedding module transforms the input tensor $(\mathbf{X}_i, \mathbf{M}_i, \mathbf{\Delta}_i)$ by independently processing each of the $D$ covariates, indexed by $d \in \{1, \cdots D\}$, per time interval $j$, preserving the covariate and time dimensions for subsequent layers. For each feature and time interval, a base embedding is generated from its observed value using a feed-forward network (FFN) for numerical inputs or an embedding look-up table for categorical ones. To explicitly handle missing data, a gating mechanism selects between the created feature embedding and a learnable, feature-specific ``missing'' embedding $z^{d}$, conditioned on the binary mask. To account for information staleness, the resulting vector is modulated by an exponential decay factor. This process yields the final time-aware embedding for $\zeta_t^{d}$,
{\small
\begin{equation}
\mathbf{\zeta}^{d}_t =
\begin{cases}
\text{Embed}_d(x^{d}_t) & \text{if } \mathbf{m}^{d}_t = 0 \\
% \text{Embed}_d(x^{d}_t) \cdot e^{-\gamma^d \delta^{d}_t} & \text{if } \mathbf{m}^{d}_t = 0 \\
z^{d} \cdot e^{-\gamma^d \delta^{d}_t} & \text{if } \mathbf{m}^{d}_t = 1,
\end{cases}
\end{equation}
}
where $\text{Embed}_d$ is the feature-specific value embedder, $\gamma^{d}$ a learnable decay rate per feature and $\delta_t^{d}$ the time elapsed since the last observation. The module outputs a tensor of shape $(B, S, D, d_{\text{emb}})$, where $S$ is the sequence length padded to the maximum length in the batch, and $d_{\text{emb}}$ is the embedding dimension. This tensor is passed to the shared factorized attention encoder.

\subsubsection{Factorized Attention Encoder}
Existing Transformer-based models for longitudinal data often create a single embedding out of all covariates for each time interval (\cite{2024transformer}, \cite{cohen2024tototimeseriesoptimized}). By summarizing feature information into a single embedding, the model loses the granular data required to learn time-varying covariate interactions. To address this, our proposed encoder introduces a factorized attention architecture that avoids this aggregation by processing the temporal and feature dimensions in an interleaved manner \cite{factorizedattention}. The encoder is a stack of identical blocks that operate on an embedding $\zeta \in \mathbb{R}^{B \times S \times D \times d_{\text{emb}}}$ with added positional encoding $P_{\text{pos}, \text{enc}}$. Within each block, two sequential attention operations are performed. First, to build a context-aware representation of each feature's trajectory, temporal attention is applied across the time dimension $S$. For input $\zeta^{l}$ to the $l$-th block, this operation consists of multi-head attention $\text{MHA}_{\text{time}}$ with a residual connection and layer normalization, 
{\small
\begin{equation}
\mathbf{\zeta}^{l}_{\text{time}} = \text{LayerNorm}(\mathbf{\zeta}^{l} + \text{MHA}_{\text{time}}(\mathbf{\zeta}^{l} + P_{\text{pos}, \text{enc}})).
\end{equation}
}
Second, to model the dynamic interactions between these contextualized features, covariate attention is applied across the covariate dimension $D$. This step uses a second multi-head attention $\text{MHA}_{\text{cov}}$, 
{\small
\begin{equation}
\mathbf{\zeta}^{l+1}_{\text{FA}} = \text{LayerNorm}(\mathbf{\zeta}^{l}_{\text{time}} + \text{MHA}_{\text{cov}}(\mathbf{\zeta}^{l}_{\text{time}})).
\end{equation}
} 
These interleaving temporal and covariate attention layers is what allows the model to learn time-varying interactions. After processing through all factorized attention blocks, the resulting history, $\mathbf{\zeta}^{l+1}_{\text{FA}}$ is distilled into a single context vector $c$. Rather than taking the last hidden state $\zeta^{\tau_i}_{\text{FA}}$, we employ a summarizing cross-attention mechanism. We introduce a learnable query $q_{\text{sum}}$, which attends to the full history to dynamically extract the most salient information from the contextual subject history. This produces the final summary context vector $c \in \mathbb{R}^{D\cdot d_{\text{emb}}}$.

\begin{table*}[t]
\centering
\small
\setlength{\tabcolsep}{1mm}

\begin{tabular}{@{}l c c l c c c ccc ccc@{}}
\toprule
\multirow{2}{*}{Dataset} &
\multirow{2}{*}{$N$ subjects} &
\multirow{2}{*}{\parbox{2.2cm}{\centering Covariates\\(num / cat)}} &
\multirow{2}{*}{Event} &
\multirow{2}{*}{\# Events} &
\multirow{2}{*}{\parbox{1.6cm}{\centering Mean\\$\tau_i$}} &
\multirow{2}{*}{\parbox{1.4cm}{\centering Censoring\\\%}} &
\multicolumn{3}{c}{Event Time $t_i$} &
\multicolumn{3}{c}{Censoring time $t_i$} \\
\cmidrule(lr){8-10}\cmidrule(lr){11-13}
 &  &  &  &  &  &  & min & max & mean & min & max & mean \\
\midrule
\multirow{1}{*}{\textsc{eICU}} & 63425 & 51 / 1 &
Sepsis & 3320 & 26.69 & 94.77 &
12.00 & 168.00 & 36.84 &
6.00 & 168.00 & 52.81 \\
\midrule
\multirow{2}{*}{\textsc{MIMIC-IV}}
  & \multirow{2}{*}{70372} & \multirow{2}{*}{53 / 2} &
  Sepsis & 27929 & 33.47 & \multirow{2}{*}{58.90} & 1.00 & 200.00 & 66.18 & \multirow{2}{*}{1.00} & \multirow{2}{*}{200.00} & \multirow{2}{*}{44.23} \\
  &  &  & Death & 1304 & 23.95 &  & 1.00 & 199.00 & 32.72 &  &  &  \\
\midrule
\multirow{2}{*}{\textsc{PBC2}}
  & \multirow{2}{*}{312} & \multirow{2}{*}{9 / 6} &
  Death & 140 & 9.93 & \multirow{2}{*}{45.83} & 1.00 & 13.75 & 4.31 & \multirow{2}{*}{3.75} & \multirow{2}{*}{14.25} & \multirow{2}{*}{8.56} \\
  &  &  & Transplant & 29 & 10.50 &  & 1.25 & 8.25 & 4.59 &  &  &  \\
\midrule
\multirow{5}{*}{\textsc{MIMIC-III}}
  & \multirow{5}{*}{2279} & \multirow{5}{*}{39 / 1} &
  Septicemia & 517 & 5.00 & \multirow{5}{*}{59.37} & 5.00 & 8.00 &  7.66 & \multirow{5}{*}{8.00} & \multirow{5}{*}{8.00} & \multirow{5}{*}{8.00} \\
  &  &  & Cereb. Hemo. & 65 & 5.00 &  & 5.00 & 8.00 & 7.48 &  &  &  \\
  &  &  & Resp. Fail. & 238 & 5.00 &  & 5.00 & 8.00 & 7.57 &  &  &  \\
  &  &  & Myo. Infarc. & 62 & 5.00 &  & 5.00 & 8.00 & 7.68 &  &  &  \\
  &  &  & Pneumonia & 44 & 5.00 &  & 5.00 & 8.00 & 7.61 &  &  &  \\
\bottomrule
\end{tabular}
\caption{Descriptive statistics of the datasets. \textsc{eICU} and \textsc{MIMIC-IV} are hourly intervals, \textsc{PBC2} intervals are three-month intervals, and \textsc{MIMIC-III} intervals are unspecified.}
\label{tab:dataset-stats}
\end{table*}

\subsubsection{Cause-Specific Hazard Subnetworks}
The final stage of {\tracer} translates the summarized subject history $c$ into a sequence of cause-specific hazard predictions for a future horizon $[\tau_{i+1}, \tau_{i+H})$. The future context tensor is fed with positional encoding, $c + P_{\text{pos}, \text{dec}}$, into a set of $K$ parallel cause-specific subnetworks (censoring not included). Each subnetwork is a small fully-connected network (FFN). Their outputs are stacked to form a logit tensor $\mathbf{f}_{\theta}(\mathbf{X}_i) \in \mathbb{R}^{B\times H\times K}$. The logits are then converted into valid discrete hazards using the multinomial normalization in Eq.~\ref{eq:hazard_softmax}.

\subsection{Loss Function}
{\tracer} is trained by minimizing the negative log-likelihood of the observed event data. To handle class imbalance, the learning objective is the weighted discrete-time competing risk likelihood \cite{bible}, where the outcome for a subject at each interval $j$ is treated as a single multinomial trial,
{
\small
\begin{equation}
\label{eq:nll_loss}
\begin{aligned}
\mathcal{L} = - \sum_{i=1}^{N} \sum_{j=\tau}^{t_i} \Biggl[ 
& \sum_{k=1}^{K} w^k y_{ij}^k \log \lambda_j^k(\mathbf{X}_i) + \\
& \left(1 - \sum_{k=1}^{K} y_{ij}^k\right) 
  \log \left( 
    1 - \sum_{k=1}^{K} \lambda_j^k(\mathbf{X}_i) 
  \right) 
\Biggr],
\end{aligned}
\end{equation}
}
where $y_{ij}^k$ is a binary indicator for subject $i$ experiencing event $k$ at interval $j$, and $w^k$ is the log-inverse frequency weight for event $k$. $\sum_{k=1}^{K} y_{ij}^k$ is 1 if any event occurs, and 0 if they survive. The first, weighted component of the loss addresses the probability of the specific event, while the second component addresses the probability of survival.

\section{Experiments}

We evaluate the performance of {\tracer} on four time-to-event datasets, including three derived from large-scale critical care databases, and one from a well-established clinical trial dataset. Dataset statistics are shown in Table~\ref{tab:dataset-stats}, and pre-processing steps for each dataset can be found in Supp. Mat. A2.

\begin{itemize}
    \item \textbf{\textsc{eICU}:} To benchmark against prior work in a single-event setting, we use a cohort of hourly data derived from the \textsc{eICU} database, following the preprocessing pipeline from Rockenshaub et al. \shortcite{rockenshaub}. The task is to predict the first occurrence of sepsis from hourly ICU measurements.
    \item \textbf{MIMIC-IV:} To evaluate {\tracer} in a competing-risk scenario with high-granularity data, we constructed a custom hourly cohort from the \textsc{MIMIC-IV} database \shortcite{mimiciv}. The task is to predict the first occurring event between two competing risks: \textit{in-ICU death} and \textit{sepsis onset} within the first 200 ICU hours.
    \item \textbf{Primary Biliary Cirrhosis:} We include the \textsc{MIMIC-III} dataset from the longitudinal study on primary biliary cirrhosis \cite{pbc2}, notably used in the evaluation of SurvTimeSurvival \shortcite{survtimesurvival}. We use a fine-grained competing-risk formulation where the goal is to predict \textit{death} versus \textit{transplant}. The dataset contains subjects with features, with time discretized to three-month intervals.
    \item \textbf{MIMIC-III:} This cohort was created from the \textsc{MIMIC-III} dataset to create a five-way competing-risk prediction \cite{mimiciiisurvival}. It comprises subjects with longitudinal measurements, with 8 follow-up intervals.
\end{itemize}
Each dataset was partitioned into a 5-fold cross-validation with a left-out test set. To leverage the longitudinal data, we employ a landmarking protocol. For each subject, a time $\tau_i$ is randomly sampled from their observation window, and the model is tasked with predicting outcomes over a fixed future horizon using the standardized history up to $\tau_i$ as input. We used AdamW as the optimizer to train {\tracer} in all experiments, with a learning rate of $1\mathrm{e}{-4}$, and weight decay of $1\mathrm{e}{-5}$. The number of factorized attention layers is chosen from $\{1,\cdots, 6\}$ with a embedding and sub-embedding size of $\{16, 32, 64\}$, and the number of attention heads is from $\{2, 4, 8\}$. The number of cause-specific subnetwork MLP layers was fixed at 2 with a GELU activation function \cite{gelu}.

\subsection{Metrics}

We compare {\tracer} against state-of-the-art baselines using two established metrics that evaluate two crucial aspects of prediction quality: discrimination and calibration.

\subsubsection{Cause-Specific Integrated Brier Score ($\text{IBS}^k$)}
The Brier score is a measure of the mean squared error between the predicted probability of an event and the actual outcome at a given time $j$ \cite{ibs}. For competing risks, the cause-specific Brier score, quantifies both the calibration and discrimination of the model for a single event type $k$,
{\small
\begin{equation*}
\text{BS}^k(j) = \frac{1}{N} \sum_{i=1}^{N} \left[ \left( I(T_i \le j, e_i=k) - \hat{F}_j^k(\mathbf{X}_i) \right)^2 \cdot W_i(j) \right]
\end{equation*}
}
where $I(t_i \le j, e_i=k)$ is the ground-truth label, $\hat{F}_j^k(X_i)$ is the model's predicted CIF for event $k$ for subject $i$ at time $j$ and $W_i(j)$ is the Inverse Probability of Censoring Weighting (IPCW) to account for censoring distribution. The IBS is the integral of the Brier score over time; a lower score represents a superior model \cite{ibs2}.

\subsubsection{Cause-Specific Time-Dependent Concordance Index}
The concordance index measures a model's ability to correctly rank subjects by risk \cite{ctd}. We use a cause-specific, time-dependent C-index that evaluates, if for a valid pair of subjects, the one who experiences event $k$ earlier is correctly assigned a higher risk score for that event,
{\small
\begin{equation*}
C_{td}^k = P(\hat{F}_{t_a}^k(\mathbf{X}_i) > \hat{F}_{t_b}^k(\mathbf{X}_j) ,|, t_a < t_b, e_i = k).
\end{equation*}
}
A value of of 1.0 denotes perfect discrimination, while 0.5 is equivalent to random chance.

\subsubsection{Calibration Analysis}
For a model to be clinically useful, high discrimination is insufficient; its predicted probabilities must be well calibrated, a requirement stressed by reporting standards such as TRIPOD+AI \cite{tripodai}. Prior work (e.g., SurvTRACE \cite{SurvTRACE}, DeepHit \cite{deephit}), however, often focuses on purely discriminative metrics like the C-index, which ignores calibration. We therefore assess calibration directly by plotting the predicted risk against observed risk. To obtain unbiased empirical estimates from censored data, we stratify predictions into bins and compute the true event rate in each using IPCW weights. In these plots, a well-calibrated model aligns with the identity line.

\section{Results}

\begin{figure}[ht]
\centering

\begin{tabular}{@{}c@{}}
  \includegraphics[width=1.0\linewidth]{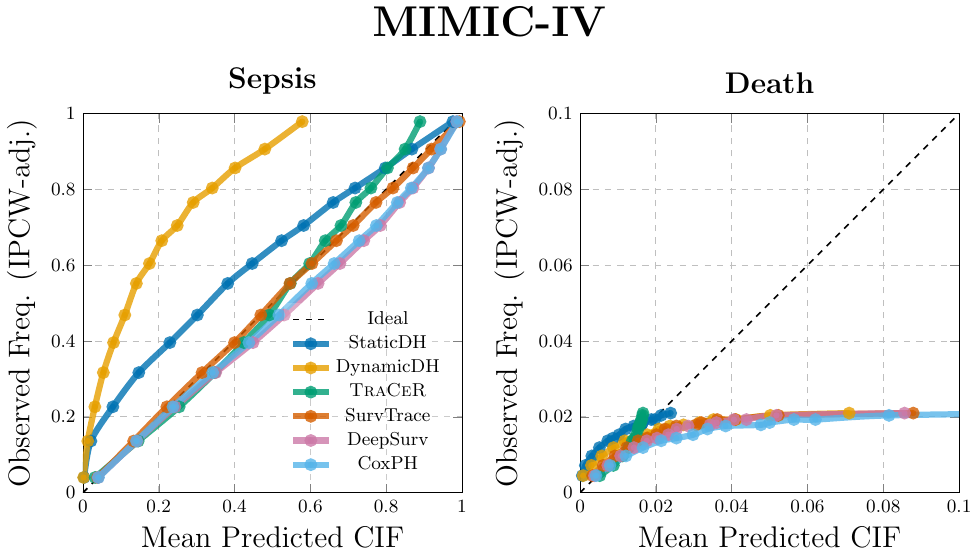} \\%[1em] % Add some vertical space
  \includegraphics[width=1.0\linewidth]{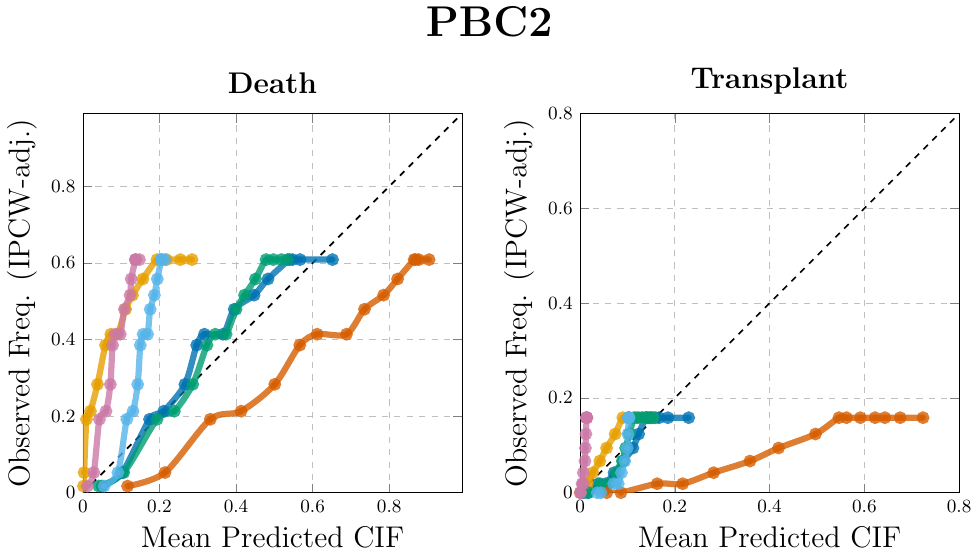}
\end{tabular}
\caption{Cause-specific calibration curves for the same test set by all methods on the \textsc{MIMIC-IV} and \textsc{PBC2} dataset. $y$-axis is the IPCW-corrected true event rate; $x$-axis is the mean predicted CIF $\hat{F}_{t_i}^k(\mathbf{X}_i)$.}
\label{fig:calibration-plot}
\end{figure}

We evaluated {\tracer} against established baselines on four clinical datasets. Our experiments were designed to assess both discriminative performance ($C_{td}$) and calibration ($\text{IBS}$, Figure~\ref{fig:calibration-plot}), with results summarized in Table~\ref{tab:ibs-table} and Table~\ref{tab:ctd-table}.

\subsection{Overall Performance}  
{\tracer} established a new state-of-the-art in predictive calibration, 
demonstrates superior $\text{IBS}$ scores across all evaluated dataset and event types (Table~\ref{tab:ibs-table}). This improvement in calibration was paired with competing discrimination scores (Table~\ref{tab:ctd-table}), where {\tracer} achieved the highest $C_{td}$ in nearly all settings, particularly on datasets with higher sampled longitudinal covariates. These successful results suggest that {\tracer}  not only correctly orders subjects by risk, but also accurately quantifies absolute risk probabilities over time. For instance, on \textsc{MIMIC-IV}, {\tracer} achieved an $\text{IBS}$ of 0.111 for sepsis prediction, a 32\% relative improvement over the next best baseline, SurvTRACE, and a 51\% relative improvement over the state-of-the-art longitudinal covariate model, DynamicDeepHit.

\subsubsection{Temporal Dynamics} The advantages of {\tracer} were most pronounced on datasets with relevant longitudinal data. On \textsc{eICU}, \textsc{MIMIC-IV} and \textsc{PBC2}, datasets with longer patient histories (Mean $\tau_i > 9$, see Table~\ref{tab:dataset-stats}), models designed to process longitudinal sequences ({\tracer}, DynamicDeepHit) significantly outperformed static baselines in $C_{td}$, with {\tracer} also achieving higher $\text{IBS}$. On \textsc{MIMIC-IV}, for sepsis prediction, {\tracer}'s $C_{td}$ of 0.816 and death prediction of 0.918, represented a 35\% and 40\% improvement over the best static model, SurvTRACE (0.603 \& 0.654), and a 7.5\% and 4\% improvement over DynamicDeepHit (0.759 \& 0.883).
Interestingly, on the \textsc{MIMIC-III} dataset, the performance gap between static and temporal models is narrow, with DeepHit achieving the highest $C_{td}$ for two of the five causes. This observation suggests that the predictive signals in this specific dataset are more heavily concentrated in the most recent subject measurement, causing prior longitudinal information to be potential harmful for modeling future hazards. This finding aligns with the results in prior work \cite{mimiciiisurvival} and highlights {\tracer}'s comparable performance on datasets where temporal dependencies are less informative.

\subsubsection{Calibration Analysis} The higher $\text{IBS}$ scores shown in Table~\ref{tab:ibs-table} are visually corroborated by the cause-specific calibration curves in Figure~\ref{fig:calibration-plot}. For call competing events in \textsc{MIMIC-IV} and \textsc{PBC2}, {\tracer}'s predicted probabilities closely follow the observed event frequencies. In contrast, prominent baselines, such as DynamicDeepHit, substantially under and overestimated risk as the prediction horizon extended. For death in \textsc{MIMIC-IV}, this behavior likely stems from the models' constraint of learning a single joint distribution over all time points and events. {\tracer}'s slight tendency to under-predict in the later intervals is likely attributed to the significant class imbalance in the dataset, which is only partially mitigated by the chosen weighing scheme. Nonetheless, {\tracer}'s calibration for sepsis onset and transplant is on par with the best baselines, and superior to all other models for death, with more calibration curves shown in Supp. Mat. B1.

\begin{table*}[!t]
  \centering
  \small
  \setlength{\tabcolsep}{1mm}

  \begin{tabular}{l c cc cc ccccc}
    \toprule
    & \multicolumn{1}{c}{\textbf{\textsc{eICU}}} & \multicolumn{2}{c}{\textbf{\textsc{MIMIC-IV}}} & \multicolumn{2}{c}{\textbf{\textsc{PBC2}}} & \multicolumn{5}{c}{\textbf{\textsc{MIMIC-III}}} \\
    
    \cmidrule(lr){2-2} \cmidrule(lr){3-4} \cmidrule(lr){5-6} \cmidrule(lr){7-11}
    
    \textbf{Approach} & Sepsis & Sepsis & Death & Death & Transplant & Septicemia & Hemorrhage & Resp. Fail. & Infarction & Pneumonia \\
    \midrule
    
    cs-Cox & 0.065$^{\dag}$ & 0.168$^{\dag}$ & 0.025$^{\dag}$ & 0.205 & 0.085$^{\dag}$ & 0.099$^{\dag}$ & 0.031$^{\dag}$ & 0.091$^{\dag}$ & 0.027$^{\dag}$ & 0.012$^{\dag}$ \\
    DeepSurv & 0.064$^{\dag}$ & 0.165$^{\dag}$ & 0.024$^{\dag}$ & 0.191 & 0.065 & 0.082$^{\dag}$ & 0.029$^{\dag}$ & 0.080$^{\dag}$ & 0.023$^{\dag}$ & 0.010$^{\dag}$ \\
    DeepHit & 0.058$^{\dag}$ & 0.176$^{\dag}$ & 0.015 & 0.184 & 0.069$^{\dag}$ & 0.062 & 0.017$^{\dag}$ & 0.051$^{\dag}$ & 0.012$^{\dag}$ & 0.006$^{\dag}$ \\
    DynamicDeepHit & 0.058$^{\dag}$ & 0.230$^{\dag}$ & 0.017$^{\dag}$ & 0.199 & 0.067$^{\dag}$ & 0.069$^{\dag}$ & 0.016$^{\dag}$ & 0.052$^{\dag}$ & 0.013$^{\dag}$ & 0.006$^{\dag}$ \\
    SurvTRACE & 0.065$^{\dag}$ & 0.164$^{\dag}$ & 0.025$^{\dag}$ & 0.199 & 0.108$^{\dag}$ & 0.088$^{\dag}$ & 0.033$^{\dag}$ & 0.088$^{\dag}$ & 0.027$^{\dag}$ & 0.011$^{\dag}$ \\
    \midrule
    
    {\tracer} w/o CET\textsuperscript{1} & 0.056 & 0.114 & 0.015 & 0.126 & 0.054 & 0.060 & \bfseries\underline{0.014} & 0.050 & 0.012$^{\dag}$ & \bfseries\underline{0.005} \\
    {\tracer} w/o FA\textsuperscript{2} & \bfseries\underline{0.055} & 0.118$^{\dag}$ & 0.015 & 0.126 & 0.091$^{\dag}$ & 0.063$^{\dag}$ & \bfseries\underline{0.014} & 0.050 & 0.012$^{\dag}$ & \bfseries\underline{0.005} \\
    {\tracer} & \bfseries\underline{0.055} & \bfseries\underline{0.111} & \bfseries\underline{0.014} & \bfseries\underline{0.113} & \bfseries\underline{0.051} & \bfseries\underline{0.061} & \bfseries\underline{0.014} & \bfseries\underline{0.049} & \bfseries\underline{0.011} & \bfseries\underline{0.005} \\

    \bottomrule
  \end{tabular}
  \caption{Mean \& standard deviation Integrated Brier Score (IBS) over a 5-fold cross-validation. Lower is better. $\dag$ indicates p $< 0.05$. \textsuperscript{1}Covariate Embedding with Time-Decay. \textsuperscript{2}Factorized Attention. Table with standard deviation in Supp. Mat. B2.}
    \label{tab:ibs-table}
\end{table*}

\begin{table*}[!t]
  \centering
  \small
  \setlength{\tabcolsep}{1mm}
  \begin{tabular}{l c cc cc ccccc}
    \toprule
    & \multicolumn{1}{c}{\textbf{\textsc{eICU}}} & \multicolumn{2}{c}{\textbf{\textsc{MIMIC-IV}}} & \multicolumn{2}{c}{\textbf{\textsc{PBC2}}} & \multicolumn{5}{c}{\textbf{\textsc{MIMIC-III}}} \\
    \cmidrule(lr){2-2} \cmidrule(lr){3-4} \cmidrule(lr){5-6} \cmidrule(lr){7-11}
    \textbf{Approach} & Sepsis & Sepsis & Death & Death & Transplant & Septicemia & Hemorrhage & Resp. Fail. & Infarction & Pneumonia \\
    \midrule
    
    cs-Cox & 0.597$^{\dag}$ & 0.580$^{\dag}$ & 0.702$^{\dag}$ & 0.592$^{\dag}$ & 0.561$^{\dag}$ & 0.871$^{\dag}$ & 0.801$^{\dag}$ & 0.783$^{\dag}$ & 0.725$^{\dag}$ & 0.765$^{\dag}$ \\
    DeepSurv & 0.634$^{\dag}$ & 0.594$^{\dag}$ & 0.498$^{\dag}$ & 0.566$^{\dag}$ & 0.581$^{\dag}$ & 0.880$^{\dag}$ & 0.749$^{\dag}$ & 0.785$^{\dag}$ & 0.732$^{\dag}$ & 0.581$^{\dag}$ \\
    DeepHit & 0.670$^{\dag}$ & 0.625$^{\dag}$ & 0.652$^{\dag}$ & 0.659$^{\dag}$ & 0.640$^{\dag}$ & 0.843$^{\dag}$ & 0.764$^{\dag}$ & \bfseries\underline{0.813} & \bfseries\underline{0.842} & 0.628$^{\dag}$ \\
    DynamicDeepHit  & 0.730$^{\dag}$ & 0.759$^{\dag}$ & 0.883$^{\dag}$ & 0.820$^{\dag}$ & 0.785$^{\dag}$ & 0.850$^{\dag}$ & 0.882 & 0.791$^{\dag}$ & 0.803$^{\dag}$ & 0.739$^{\dag}$ \\
    SurvTRACE       & 0.649$^{\dag}$ & 0.603$^{\dag}$ & 0.654$^{\dag}$ & 0.552$^{\dag}$ & 0.603$^{\dag}$ & \bfseries\underline{0.888} & 0.805$^{\dag}$ & 0.790$^{\dag}$ & 0.714$^{\dag}$ & 0.803 \\
    \midrule
    
    {\tracer} w/o CET\textsuperscript{1}   & 0.857$^{\dag}$ & 0.808$^{\dag}$ & 0.898 & 0.889 & 0.879 & 0.849$^{\dag}$ & \bfseries\underline{0.887} & 0.807 & 0.739$^{\dag}$ & \bfseries\underline{0.832} \\
    {\tracer} w/o FA\textsuperscript{2}    & 0.864$^{\dag}$ & 0.794$^{\dag}$ & 0.826$^{\dag}$ & 0.878 & 0.851 & 0.846$^{\dag}$ & 0.839$^{\dag}$ & 0.809 & 0.777$^{\dag}$ & 0.731$^{\dag}$ \\
    {\tracer} & \bfseries\underline{0.870} & \bfseries\underline{0.816} & \bfseries\underline{0.918} & \bfseries\underline{0.891} & \bfseries\underline{0.901} & 0.850$^{\dag}$ & 0.863$^{\dag}$ & 0.810 & 0.728$^{\dag}$ & 0.782$^{\dag}$ \\

    \bottomrule
  \end{tabular}
  \caption{Mean \& standard deviation Concordance Index ($C_{td}$) over a 5-fold cross-validation. Higher is better. $\dag$ indicates p $< 0.05$. \textsuperscript{1}Covariate Embedding with Time-Decay. \textsuperscript{2}Factorized Attention. Table with standard deviation in Supp. Mat. B2.}
  \label{tab:ctd-table}
\end{table*}

\subsubsection{Ablations} Our ablation studies confirm the contribution of {\tracer}'s core architectural components. The removal of factorized attention ({\tracer} \text{w/o FA}) resulted in a significant performance decrease across both metrics, particularly on \textsc{MIMIC-IV}. This result strengthens our hypothesis that the explicit attendance of temporal and covariate interactions is beneficial for modeling longitudinal data, and an advantage over global-embedding methods. Similarly, removing the covariate embedding ({\tracer} \text{w/o CET}) slightly harmed performance, indicating a benefit to informing the model about time gaps between measurements. On \textsc{MIMIC-III}, all variants of {\tracer} performed similarly, further supporting the notion that the dataset's outcomes are largely driven by the latest measurement, rendering the enhanced temporal components less impactful, if not harmful. Further model ablations testing summarization and positional encoding strategies can be found in Supp. Mat. B3.

\section{Discussion}
Our results demonstrate that {\tracer} establishes a new state-of-the-art in survival analysis, not only in terms of discriminative ability, but, more critically, in its generation of well-calibrated risk predictions. We attribute {\tracer}'s improved performance to its factorized attention backbone. Unlike RNN-based models, such as DynamicDeepHit, that process subject information sequentially, or prior Transformer approaches that decouple temporal and feature interactions, {\tracer} iteratively builds a summary of known covariate history by alternating attention across time and covariate axis. The temporal attention initially learns the trajectory of each covariate independently. The subsequent covariate attention then models the dynamic influence between these contextualized trajectories. This allows the model to capture more complex time-varying relationships.
{\tracer}'s improved calibration also stems from its direct modeling of the cause-specific hazard function, a key difference from PMF-based methods like the DeepHit family of models. The negative log-likelihood loss on the discrete hazard (Eq.~\ref{eq:nll_loss}) is inherently modular, in that the loss for a given interval $j$ is decoupled from future intervals. In contrast, training on a global PMF requires normalizing over all future time points and event types simultaneously. While PMF models achieve high discrimination, notably with the use of a ranking loss, this can come at the cost of probabilistic accuracy \cite{deephitbadcalibration}. Our approach, by directly optimizing the likelihood of the conditional event probability at each step, provides a more stable learning target that aligns better with the goal of accurate risk estimation.

Our experimental design employed a landmarking approach to ensure fair comparison with existing static models. While effective for benchmarking, we acknowledge this does not fully exploit the potential for dynamic predictions \cite{landmarking}. However, unlike static models, {\tracer} is not constrained by the architecture, as causal-masked attention (to prevent data leakage) and the negative likelihood loss are inherently suited for dynamic predictions. Furthermore, while factorized attention is more computationally intensive than RNNs, it is more parameter efficient than a naive Transformer that would flatten the time and covariate dimensions, avoiding a quadratic attention complexity.

Finally, our work aligns with the growing consensus in clinical ML, such as the guidelines of TRIPOD+AI \shortcite{tripodai}, that robust calibration analysis is required for trustworthy clinical AI. Beyond predictive accuracy, the factorized attention offers powerful interpretability. As demonstrated in Supp. Mat. B4, the model learns that earlier time points can be more predictive than the most recent data, assigning them higher attention weight. This research opens several avenues for future work. A natural extension beyond dynamic survival analysis is multi-state modeling, where {\tracer} could be adapted to learn transition-specific hazards between several states (e.g., stable $\rightarrow$ sepsis $\rightarrow$ death). Future work can also explore methods to model informative censoring and investigate {\tracer}'s generalization across harmonized clinical datasets relating to our \textsc{eICU} cohort.

\bibliography{aaai2026}

% \input{LaTeX/ReproducibilityChecklist}

\end{document}